\documentclass{article}
\usepackage{ijcai22}

\usepackage{times}
\usepackage{soul}
\usepackage{url}
\usepackage[hidelinks]{hyperref}
\usepackage[utf8]{inputenc}
\usepackage[small]{caption}
\usepackage{graphicx}
\usepackage{amsmath}
\usepackage{amsthm}
\usepackage{booktabs}
\usepackage{algorithm}
\usepackage{algorithmic}
\usepackage{amssymb}

\usepackage[short]{optidef}
\usepackage[table,xcdraw]{xcolor}

\theoremstyle{definition}
\newtheorem{definition}{Definition}[section]

\title{An Attention Model for the Formation of Collectives in Real-World Domains}

\author{
Adrià Fenoy$^{1,2}$
\and
Filippo Bistaffa$^2$\and
Alessandro Farinelli$^{1}$
\affiliations
$^1$University of Verona\\
$^2$IIIA-CSIC
\emails
adria.fenoybarcelo@univr.it,
filippo.bistaffa@iiia.csic.es,
alessandro.farinelli@univr.it
}

\usepackage{glossaries}
\newacronym{vrp}{VRP}{\emph{vehicle routing problem}}
\newacronym{ilp}{ILP}{\emph{integer linear program}}
\newacronym{milp}{MILP}{\emph{mixed-integer linear program}}
\newacronym{wsp}{WSP}{\emph{weighted set packing}}
\newacronym{AI}{AI}{Artificial Intelligence}
\newacronym{mcts}{MCTS}{\emph{Monte Carlo tree search}}

\begin{document}

\maketitle

\begin{abstract}
We consider the problem of forming collectives of agents for real-world applications aligned with \emph{Sustainable Development Goals} (e.g., shared mobility, cooperative learning).
We propose a general approach for the formation of collectives based on a novel combination of an \emph{attention model} and an \gls{ilp}.
In more detail, we propose an attention encoder-decoder model that transforms a collective formation instance to a \acrlong{wsp} problem, which is then solved by an \gls{ilp}.
Results on two real-world domains (i.e., ridesharing and team formation for cooperative learning) show that our approach provides solutions that are comparable (in terms of quality) to the ones produced by state-of-the-art approaches specific to each domain. Moreover, our solution outperforms the most recent general approach for forming collectives based on \emph{Monte Carlo tree search}.
\end{abstract}

\section{Introduction}

In recent years, more and more scenarios require \emph{Collective Intelligence} solutions enabling novel ways of social production, promoting innovation and encouraging the exchange of ideas \cite{eu}. Such new forms of collaborative consumption and production rely on a common and fundamental task, i.e., \emph{the formation of collectives}.
This task plays a crucial role in a general class of real-world applications aligned with the \emph{UN Sustainable Development Goals}, which enable agents to complete tasks and achieve benefits by means of cooperation. 

As an example of this class of domains, in \emph{ridesharing} commuters can form groups and travel together with the objective of reducing transportation costs, mitigate pollutant emissions and alleviate the traffic congestion in urban environments \cite{bistaffa2019computational,alonso2017demand}.
Another prominent example can be found in modern educational institutions that aim at implementing cooperative and active learning techniques, which engage students in \emph{teams} to participate in all learning activities in the classrooms.
Indeed, as recently shown in \cite{andrejczuk2019synergistic} collective formation approaches can improve the overall performance of the students by grouping them in teams that maximize the synergies among members.

Due to the inherent complexity and specificity of each application domain, researchers usually tackle the formation of collectives by designing very specific sub-optimal approaches that can solve the associated large-scale optimization problem in a feasible runtime.
Unfortunately, one domain-specific approach usually cannot be applied in a different scenario.

In contrast, in this paper, we propose a novel, general approach for the formation of collectives that is based on two fundamental steps.
First, we apply deep reinforcement learning techniques to train an attention encoder-decoder model, with the objective of automatically generating a set of \emph{promising collectives} on the basis of the structure of the considered scenario. 
In the second step, we compile a \gls{wsp} instance that, by only taking into account the promising candidates generated in the first step, can be solved by off-the-shelf \gls{ilp} solvers in a manageable time budget.
Thus, our approach does not require to manually specify any domain-specific knowledge, in contrast with the above-mentioned sub-optimal state-of-the-art approaches.
Furthermore, by only considering a set of promising candidates rather than the entire set of possible collectives\footnote{The number of possible collectives grows exponentially with the number of agents, hence it is not manageable in real-world applications that involve more than a few tens of agents}, we reduce the complexity of the original problem by several orders of magnitude without sacrificing the quality of the final solution that we produce.

As such, this paper advances the state-of-the-art as follows:

\begin{itemize}
    \item We propose a general approach for the formation of collectives in real-world domains based on the novel combination of an attention model and \gls{wsp}.
    \item We proposed a novel training procedure for our attention model based on \emph{Maximum Entropy Reinforcement Learning}. In contrast to previous approaches which use attention based models for optimization \cite{kool2018attention}, our solution achieves a wide variety of promising candidates. Such variety is a key feature that allows the \gls{ilp} solver to group collectives of high value.
    \item We evaluate our approach in two real-world scenarios (i.e., ridesharing and team formation for cooperative learning) by comparing it with state-of-the-art approaches specific to each domain. Our results show that our approach can produce, in some settings, solutions of comparable quality without requiring any domain-specific knowledge. Moreover, we compare our approach with the most recent general approach for forming collectives based on \emph{Monte Carlo tree search} \cite{wu2020monte}, showing that our solutions clearly outperform (in terms of quality) the ones computed by the counterpart.
\end{itemize}

\section{Background \& Related Work}


\subsection{Formation of Collectives}\label{sec:backcf}

The general problem of forming collectives of agents has been deeply studied from many different perspectives in the scientific literature.
In this paper we specifically focus on the optimization problem \cite{cerquides2014tutorial} of computing the set of non-overlapping collectives (i.e., subsets) of agents belonging to a universal set $A$.
Each collective is associated to a value provided by a domain-specific utility function, e.g., the reduction in terms of cost or CO\textsubscript{2} emissions associated to the arrangement of a shared trip \cite{bistaffa2019computational} or the improvement thanks to cooperation within a team of students \cite{andrejczuk2019synergistic}.
Thus, our objective is to maximize the sum of the values associated with each formed collective.

By and large, the formation of collectives requires solving a \emph{set partitioning} problem \cite{lin1975corporate} or, equivalently, a \emph{coalition structure generation} problem \cite{michalak2016hybrid}.
A wealth of approaches have been proposed to tackle the formation of collectives from this perspective \cite{rahwan2015coalition}.
General algorithms \cite{changder2020odss,michalak2016hybrid} usually make no assumptions on the structure of the utility function, which is treated as a \emph{black-box} oracle.
Unfortunately, the mere act of providing the input to the solution algorithm (without even considering the runtime of the algorithm itself) requires enumerating a number of values that grows exponentially with the number of agents and hence becomes quickly not manageable. 
Notice that in many application domains it is possible to exploit domain specific constraints that limit the number of possible collectives. For example, cardinality constraints that limits the maximum size of the collectives to $k$ naturally arise in ridesharing and team formation. However, even considering such a constraint the number of collectives is $O(|A|^k)$, which can require \emph{hours} to enumerate when the computation of the utility function is complex.

To overcome this limitation, the formation of collectives in real-world domains is usually tackled by means of sub-optimal approaches that trade generality for scalability, i.e., that exploit the specific structure of the considered domain to compute good-quality solutions in a feasible amount of time.
For instance, \cite{bistaffa2019computational} proposed a solution algorithm for large-scale ridesharing that, by heavily relying on the greedy nature of the domain, is capable of computing solutions of very good quality for hundreds of agents within one minute.
Unfortunately, this approach cannot be applied in collective formation domains that are not characterized by such a greedy nature, e.g., the team formation domain discussed in \cite{andrejczuk2019synergistic}.
Here, the authors proposed a local-search algorithm that, once again, heavily relies on the structure of the problem and the considered dataset.

Against this background, in this paper we propose an approach for the formation of collectives that does not require to manually specify any domain-specific knowledge but aims at learning such structure by applying deep reinforcement learning techniques.

\subsection{Machine Learning for Optimization}

The use of machine learning techniques to solve combinatorial optimization problems is a recent yet very active topic that has received a lot of attention during the last years.
According to \cite{bengio2020machine}, machine learning can contribute to the optimization field in a twofold way: i) replace some heavy computations by building fast approximations and ii) improve the optimization approach by learning domain-specific structure.

In the first and most common case, machine learning is employed to train a model so as to ``imitate'' the behavior of the original algorithm in terms of solution quality, but being much faster.
Relevant examples are the work of \cite{kool2018attention}, which proposes a machine learning approach based on an attention model to solve several variations of the \gls{vrp} or the work of \cite{nair2021solving}, which proposes a deep learning model to solve \glspl{milp}.

The second direction mentioned by \cite{bengio2020machine} is more aligned with our work.
Indeed, we consider a scenario where the algorithmic decisions (in our case, deciding what are the best collectives to form) rely on highly specific hard coded knowledge that is difficult and costly to acquire. 
Thus, our goal is to employ machine learning to automatically exploit the structure of the domain and learn the best performing behavior (i.e., a policy) to guide the formation of collectives.

In the following section we discuss our proposed approach that aims at achieving this objective.

\section{Our Solution Approach}
\label{sec:ourapproach}

Given a pool of $n$ agents $A = \{a_1, a_2, \ldots, a_n\}$, we tackle the formation of collectives as the problem of computing the best set $\mathcal{S}$ of non-overlapping subsets of $A$ which optimizes the sum of the values associated to each subset $S\in\mathcal{S}$ by a utility function $f : \mathcal{F}(A) \to \mathbb{R}$. Such utility function maps every collective in the feasible set\footnote{Depending on the considered domain, such a set of feasible collectives can be the entire set of subsets of $A$ or, for example, the set of all collectives that satisfy a given constraint (e.g., a cardinality constraint).} of collectives $\mathcal{F}(A)$ to a real number.
This problem can be naturally formulated as an \gls{ilp}:
\begin{equation}\label{ILP}
\begin{aligned}
    \mbox{maximize} \enspace & \sum_{S \in \mathcal{F}(A)} f(S) \cdot x_S,\\
    \mbox{subject to} \enspace & \sum_{S \in \mathcal{F}(A)} b_{i,S} \cdot x_S \leq 1, \quad \forall a_i \in A,
\end{aligned}
\end{equation}
where $x_S$ is a binary decision variable that encodes whether collective $S$ is in the set $\mathcal{S}$ and $b_{i,S}$ is a binary value that encodes whether agent $a_i \in A$ belongs to the collective $S$.

The problem in \eqref{ILP} can be easily recognized as a \gls{wsp} and, for small-scale instances (i.e., $A$ with less than a couple of tens of agents), directly solved as an \gls{ilp} by means of off-the-shelf solvers.
On the other hand, in real-world scenarios the \emph{generation} of such an \gls{ilp} (let alone its solution) can require hours of computation, due to the necessity of enumerating all feasible collectives. This complexity is further increased in scenarios where determining each value $f(S)$ requires a significant computational effort \cite{andrejczuk2019synergistic}.

On the other hand, usually, by exploiting the inherent structure of the domain, an expert might propose a reduced set of \emph{promising} collectives $\mathcal{R}(A)$, from which a sub-optimal solution of high quality can be obtained.
Following this approach, in this paper we propose an attention-based model that learns this structure to generate an  \emph{\gls{ilp} of manageable size}, i.e.,  
\begin{equation}\label{rILP}
\begin{aligned}
    \mbox{maximize} \enspace & \sum_{S \in \mathcal{R}(A)} f(S) \cdot x_S,\\
    \mbox{subject to} \enspace & \sum_{S \in \mathcal{R}(A)} b_{i,S} \cdot x_S \leq 1, \quad  \forall a_i \in A.
\end{aligned}
\end{equation}

\begin{figure}[ht]
    \centering
    \includegraphics[width=\columnwidth]{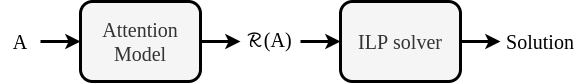}
    \caption{Proposed approach for the formation of collectives. The attention model generates a reduced set of collectives from which an \gls{ilp} solver obtains the solution.}
    \label{general_approach}
\end{figure}

In Figure \ref{general_approach} we provide an overall scheme that illustrates our approach for the formation of collectives.

Following a standard practice \cite{bistaffa2019computational}, we assume that the entire approach depicted in Figure \ref{general_approach} is provided with a time budget $t$ in which a solution has to be computed.
Notice that, since our approach comprises two phases, we need to distribute such a time budget for each of the two steps, hence we assign a runtime of  $k \cdot t$ to the first phase (i.e., generation of the \gls{wsp} instance via the attention model) and the remaining part $(1 - k) \cdot t$ to the solution of such instance by the \gls{ilp} solver.

\subsection{Attention Model}


Our attention model can be considered as a decision-making process, where collectives are built incrementally by selecting elements from the set of agents $A$ and adding them to the collective. Our model receives the set $A$ as a list of $d_x$ dimensional vectors, where $d_x$ is the number of features, and a binary encoding of a collective $S = \{b_{1,S}, b_{2,S}, \ldots, b_{n,S}\}$, where $b_{i,S}$ are binary values determining whether the respective agents $a_i$ are in the collective or not. Therefore, the state of the problem at each step is represented by the tuple $s = (A, S)$

\begin{figure}[ht]
    \centering
    \includegraphics[width=0.8\columnwidth]{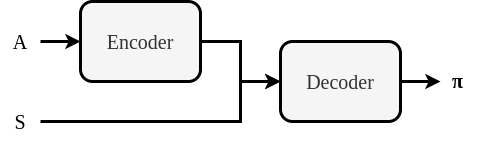}
    \caption{The encoder-decoder approach computes the probability $\pi_{\boldsymbol{\theta}}$ for each agent in $A$ to be added to the collective $S$.}
    \label{encoder_decoder}
\end{figure}

Given a state $s$, we design an attention-based encoder-decoder model based on the one proposed by \cite{vaswani2017attention} which defines a stochastic policy $\pi_{\boldsymbol{\theta}}(s)$ parameterized by $\boldsymbol{\theta}$, determining the probability for each element in the pool of agents $A$ to be included in the collective $S$. The encoder produces an embedding for each element in the pool. Then, as illustrated in Figure \ref{encoder_decoder}, the decoder receives the embedding and the collective in order to compute the probabilities.

\subsubsection{Encoder}

\begin{figure}[ht]
    \centering
    \includegraphics[width=0.5\columnwidth]{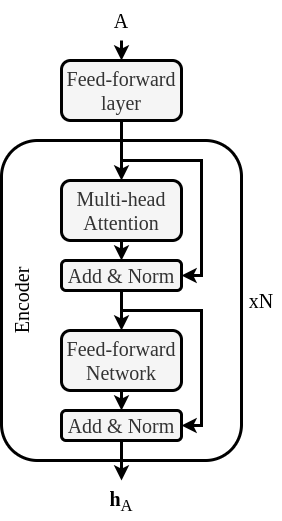}
    \caption{Encoder architecture.}
    \label{encoder}
\end{figure}

Our encoder is similar to the one in the architecture discussed by \cite{vaswani2017attention}.
In contrast with the original model, we omit positional encoding, since the order of the elements in the pool of agents is not relevant for the formation of collectives. Nevertheless, we use an input feed-forward layer to encode elements in the pool of agents $A$ from its $d_x$ dimensional feature representation to a $d_h$ dimensional embedding before main attention blocks.
To get the encoded representation of the pool of agents $\boldsymbol{h}_{A}$, the input embeddings are updated using $N$ attention blocks depicted in Figure \ref{encoder}, each one consisting of two sub-layers: a multi-head self-attention and a feed-forward layer. Each sub-layer adds a residual connection \cite{he2016deep} and performs layer normalization \cite{ba2016layer} on its outputs, i.e., $\text{LayerNorm}(x + \text{sub-layer}(x))$. To facilitate residual connections, all sub-layers in the encoder use the same dimensionality $d_h$.

\subsubsection{Decoder}

In order to compute the probabilities $\pi_{\boldsymbol{\theta}}(s)$, the decoder performs attention between the encoded pool $\boldsymbol{h}_{A} = \{ \boldsymbol{h}_1, \boldsymbol{h}_2, ..., \boldsymbol{h}_n\}$ and an encoding of the collective $\boldsymbol{h}_{S}$. To obtain this encoding, the following reduction is applied on the encoded pool:
\begin{equation}
    \label{encodedS}
    \boldsymbol{h}_{S} = \frac{\sum_{i} b_{i, S}  \cdot \boldsymbol{h}_{i}}{\sum_{i} b_{i, S}}.
\end{equation}
At the initial state the collective is empty, which means that ${\sum_{i} b_{i, S}} = 0$, in that case we use a $d_h$ dimensional learnable parameter $\boldsymbol{v}$ as a placeholder, $\boldsymbol{h}_S = \boldsymbol{v}$.

Finally, to obtain the probabilities $\pi_{\boldsymbol{\theta}}(s)$, the decoder performs two last attention steps. The first one computes attention between ${\boldsymbol{h}_{A}}$ and $\boldsymbol{h}_{S}$ to obtain a combined encoding $\boldsymbol{h}'$. The second step computes the compatibility $u_i$ between $\boldsymbol{h}_S$ and $\boldsymbol{h}'$,
\begin{equation}
    u_{i} = \frac{(\boldsymbol{h}_S W^q ) (\boldsymbol{h}'_{i} W^k)^T} {\sqrt{d_h}} ,
\end{equation}
where $W^q$ and $W^k$ are two learnable linear transformations and the output is scaled with a factor of $\frac{1}{\sqrt{d_h}}$. The compatibility is normalized by applying a softmax in order to obtain the probability of each agent being added to a collective $S$:
\begin{equation}
    \pi_{\boldsymbol{\theta}, i}(s) = \frac{e^{\gamma \tanh{u_{i}}}} {\sum_{j} e^{\gamma \tanh{u_{j}}}} ,
\end{equation}
where the function $\gamma \tanh u_i$ is applied onto the compatibilities so as to control the exploration of the model by adjusting parameter $\gamma$.

\subsubsection{Maximum Entropy Policy Gradient}

In the previous section we defined the attention model for computing the probabilities $\pi_{\boldsymbol{\theta}}(s)$ for the formation of collectives modeled as a decision making process.
In this section, we elaborate on how we optimize the parameters $\boldsymbol{\theta}$ for this task.

In the context of such a discussion, it is important to recall our ultimate goal: forming a set of collectives $\mathcal{R}(A)$ from which, by means of an \gls{ilp} solver, it can be obtained a solution of good quality for a particular instance of a collective formation problem.
For our approach to be effective, we have to guarantee (i) that $\mathcal{R}(A)$ contains collectives associated to high utility values by the function $f$, but also that (ii) such set contains a sufficient number of diverse collectives.
Such a diversity is fundamental because, due to the presence of the non-overlapping constraint in \eqref{ILP}, the optimal solution is likely to contain, not only collectives with the highest possible value, but also collectives of lower value. 
Henceforth, providing a sufficient number of alternatives to the \gls{ilp} solver is crucial to achieve a final solution of good quality.
Along these lines, we define the loss:
\begin{equation}
    \mathcal{L}(\theta|s) = \mathbb{E}_{\pi_{\boldsymbol{\theta}}(S|s)} \left[ f(S) \right] + \tau \mathcal{H}(\pi_{\boldsymbol{\theta}}(s)),
    \label{loss}
\end{equation}
where $\mathcal{H}(\pi_{\boldsymbol{\theta}}(s))$ is the entropy of the model at state $s$ and $\tau$ is a temperature parameter.
Optimizing the first term in \eqref{loss} produces a policy which builds collectives of high utility. In addition, we consider a second entropy term to the loss, whose objective is to foster diversity.

Similar to \cite{kool2018attention} we optimize our model by gradient descent with the well-known REINFORCE algorithm \cite{williams1992simple}. However, in contrast to such work, we introduce an additional entropy term:
\begin{equation}
    \begin{aligned}
        \nabla_{\boldsymbol{\theta}} \mathcal{L} &= \mathbb{E}_{\pi_{\boldsymbol{\theta}}(S | s)}  \left[ \left( f(S) - b(s) \right) \nabla_{\boldsymbol{\theta}} \log \pi_{\boldsymbol{\theta}}(S | s) \right] + \\ &+ \tau \nabla_{\boldsymbol{\theta}} \mathcal{H}(\pi_{\boldsymbol{\theta}}(s)) ,
    \end{aligned}
\end{equation}
where $b(s)$ is a baseline to reduce the variance of the gradient. Popular choices for the baseline are using an exponential moving average or training a critic to estimate value function given an state $s$. While the first one does not provide a baseline for a particular state $s$, the second one produces a complex training setup with two networks to optimize simultaneously. Therefore, we opted to compute the value with a rollout baseline, which estimates the value by performing rollout from a given state with the best policy obtained so far.

\begin{algorithm}[ht]
    \caption{REINFORCE with Rollout Baseline}
    \label{REINFORCE}
    \textbf{Input}: number of epochs $E$, number of iterations per epoch $I$, batch size $B$, significance $\alpha$\\
    \textbf{Output}: trained model parameters $\boldsymbol{\theta}$ \\ \vspace{-12pt}
    \begin{algorithmic}[1]
        \STATE Init $\boldsymbol{\theta}$
        \FOR{$epoch = 1 , \dots , E$}
            \FOR{$iter = 1 , \dots , I$}
                \STATE $s_i \gets \text{randomState}()$ \hspace{8pt} $\forall i \in \{ 1, \dots, B\}$
                \STATE $S_i \gets \text{rollout}(s_i, \boldsymbol{\theta})$ \hspace{8pt} $\forall i \in \{ 1, \dots, B\}$
                \STATE $S_i^{BL} \gets \text{rollout}(s_i, \boldsymbol{\theta}_{BL})$ \hspace{8pt} $\forall i \in \{ 1, \dots, B\}$
                \STATE $\nabla \mathcal{L} \gets \sum_{i=1}^B \left( f(S_i) - f(S_i^{BL}) \right) \nabla_{\boldsymbol{\theta}} \log \pi_{\boldsymbol{\theta}}(S_i | s_i)$
                \STATE $\nabla \mathcal{L}_{\mathcal{H}} \gets \tau \sum_{i=1}^B \nabla_{\boldsymbol{\theta}} \mathcal{H}(\pi_{\boldsymbol{\theta}}(s_i)) $
                \STATE $\boldsymbol{\theta} \gets \text{Adam}(\boldsymbol{\theta}, \nabla \mathcal{L} + \nabla \mathcal{L}_{\mathcal{H}})$
            \ENDFOR
            \IF{OneSidedPairedTTest($\pi_{\boldsymbol{\theta}}$, $\pi_{\boldsymbol{\theta}_{BL}}$) $\leq \alpha$}
                \STATE $\boldsymbol{\theta}_{BL} \gets \boldsymbol{\theta}$
            \ENDIF
        \ENDFOR
        \STATE \textbf{return} $\boldsymbol{\theta}$
    \end{algorithmic}
\end{algorithm}

After the loss is computed, the model parameters $\boldsymbol{\theta}$ are updated using an Adam optimizer \cite{kingma2014adam}. At the end of each epoch, the model and the baseline are evaluated by performing a complete rollout over several examples. Then, the models are compared by means of a paired T-test. In the case that the model outperforms the baseline, the last one is updated with the model parameters. The full training procedure is detailed in Algorithm \ref{REINFORCE}.

\section{Experimental Evaluation}
\label{sec:exp}

The main objective of our experimental evaluation is to assess the performance of our general collective formation approach in two structurally different real-world scenarios. On the one hand, we consider the ridesharing scenario discussed in \cite{bistaffa2019computational}, where, as the authors show, an algorithm characterized by a strongly greedy component can produce solutions close to the optimal for hundreds of agents within one minute.
On the other hand, we consider the team formation scenario discussed in \cite{andrejczuk2019synergistic}, in which greedy approaches cannot be used due to the presence of domain-specific constraints. 

\subsection{Application Domains}
\label{sec:case}

The ridesharing problem discussed in \cite{bistaffa2019computational} takes place in a map of zones $\mathcal{Z} = \{z_1, z_2, \ldots, z_m\}$. An instance of this problem involves a pool of agents $A = \{a_1, a_2, \ldots, a_n\}$, where each agent wants to travel from an origin to a destination, formally $a_i \in \mathcal{Z} \times \mathcal{Z}$. In the ridesharing domain we consider collectives with cardinality $1 \leq |S| \leq 5$ to reflect the usual capacity of cars. Each collective has an associated value assigned by a utility function $f(S)$ which represents the quality of service (e.g., the delay experienced by the users) and environmental benefits (e.g., the reduction of pollutant emissions or traffic) for the agents inside the collective.

The team formation problem discussed in \cite{andrejczuk2019synergistic} consists of a set of students $A = \{a_1, a_2, \ldots, a_n\}$, which are assigned a task that has to be solved cooperatively in a team. In all our experiments, we consider the ``English'' task.
Each student is represented by a tuple $(g, \boldsymbol{p}, \boldsymbol{l})$, where $g$ is a binary value indicating the gender, $\boldsymbol{p}$ is a vector with four personality traits, evaluated in the range $[-1, 1]$, and $\boldsymbol{l}$ is a vector with seven competence levels in the range $[0, 1]$. Each task involves covering different competences that have to be covered by a student in the team.

Notice that, since the goal of team formation is to obtained a balanced set of teams so as to foster cooperation and inclusiveness, the authors of \cite{andrejczuk2019synergistic} originally defined the team formation problem as the maximization of a \emph{Nash product}, which is then transformed into a linear optimization problem by considering the sum of the logarithms of the utility values of the teams. Here we adopt the same transformation, i.e., we consider the linearized formalization of team formation.

\subsection{Baselines}
\label{sec:baselines}

To evaluate the performance of our approach in each of the above-mentioned domains, we employ the state-of-the-art approaches proposed in \cite{bistaffa2019computational} and \cite{andrejczuk2019synergistic}, which we denote as PG\textsuperscript{2} and SynTeam, respectively.
For both approaches we use the parameters specified by the authors.

We remark that, as already mentioned in Section \ref{sec:backcf}, these approaches already achieve close-to-optimal performance in their respective domains, hence our goal here is \emph{not} to claim an improvement over these domain-specific solutions.
We also remark that neither PG\textsuperscript{2} nor SynTeam can be used outside of the domain in which they were originally designed.
Thus, our goal is to show that our general approach can provide a performance comparable to these approaches without being restricted to any specific application domain.

Additionally, we compare our approach to the \gls{mcts} algorithm presented in \cite{wu2020monte}, which, to the best of our knowledge, is the most recent general approach for the formation of collectives. Notice that such approach uses a greedy rollout policy based on selecting collectives with the best value increment at each step. As already mentioned above, such a greedy policy can not be directly used in the team formation domain, which prevents the approach from \cite{wu2020monte} from finding any feasible solution in this case.

For this reason, we decided to consider a second version of \gls{mcts} that employs an heuristic which prevents choosing actions which might lead to an unfeasible collective during rollout.
For the sake of completeness, we also consider a standard \gls{mcts} that employs a random rollout, i.e., that selects actions from a uniform distribution.
These three \gls{mcts} approaches are referred to as G-\gls{mcts} (greedy), A-\gls{mcts} (i.e., adapted) and R-\gls{mcts} (i.e., random), respectively.

\subsection{Methodology}

We test the above-mentioned algorithms using real-world datasets obtained by the authors of the articles of the two considered case studies.
Specifically, we consider $50$ problem instances for ridesharing and $20$ problem instances for team formation.
For each instance, we run each algorithm using $50$ different seeds (i.e., $0,\ldots,49$) and we compute the ratio between the average of the obtained solution values and the value of the optimal solution, i.e., the one obtained by solving \eqref{ILP} to optimality. In the experiments we scale to sizes for which we cannot compute the optimal, thus we used the state-of-the-art approach as a reference for these sizes.
We then report the average over all instances of such optimality ratios.
We do not report standard deviations since in all cases is $< 0.02$.

\paragraph{Training \& Evaluation}

The runtime and the hardware employed for training and evaluation are the following:

\begin{itemize}
    \item Training times may vary depending on the size of the training instances, but in general it takes between $12$ and $24$ hours on standard hardware (NVIDIA RTX 2080 Ti GPU).
    \item For evaluation we consider a total time budget of $60$ seconds. We employ IRACE \cite{lopez2016irace}, a widely used software for tuning algorithmic parameters. Specifically we use it to determine the portion of the total time budget devoted to each part of our algorithm, as discussed in Section \ref{sec:ourapproach}. The optimal portion of time budget devoted to the generation of promising candidates is $50$ seconds. We observed that our model is capable of generating tens of thousands of collectives during this time budget on a Tesla Volta V100 PCIe GPU.
\end{itemize}
Our attention model is implemented in PyTorch. We employ CPLEX 20.1.0.0 as an \gls{ilp} solver.

\paragraph{Hyperparameters}

We initialize the model parameters with $Uniform(-1/\sqrt{d}, 1/\sqrt{d})$, where $d$ is the input size. For the attention mechanism we use $8$ heads and $d = 256$, whereas for the feed-forward layers we use $d = 512$. The encoder is composed of $3$ attention blocks. We train the models during $100$ epochs consisting of $400$ batches with $256$ instances each. For evaluation we use $100$ batches with the same number of instances. For optimization of the model parameters we use a learning rate of $10^{-4}$ and a significance of $\alpha = 0.05$ for the one-sided paired T-test. For the purposes of this work, we did not perform an exhaustive parameter optimization. We recommend doing further parameter search in order to obtain more refined models.

\subsection{Results}

\begin{figure}[t]
    \centering
    \includegraphics[width=0.9\columnwidth]{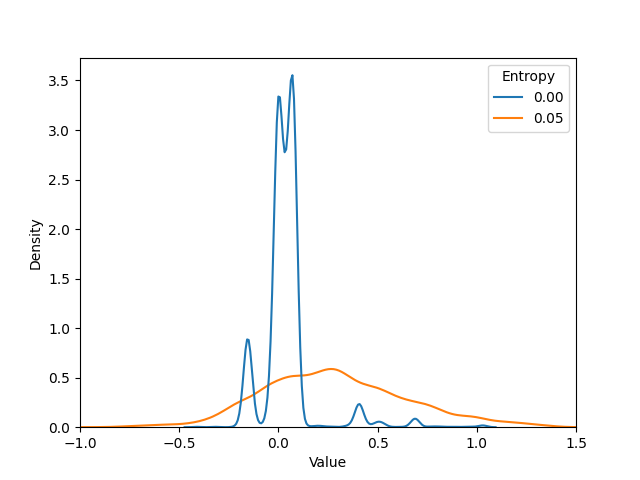}
    \caption{Probability density of the values of collectives generated by our attention model employing $\tau = 0.00$ and $\tau = 0.05$ for an example ridesharing problem instance.}
    \label{hist}
\end{figure}

One of the most important findings we report from our experimental evaluation is that a higher entropy term produces more variety in the set of candidate collectives.
Indeed, Figure \ref{hist} shows the distribution of values of the collectives generated by our model for one example ridesharing instance.
It is clear that the model with the higher entropy term results in a higher diversity, while also producing collectives characterized by slightly higher utility values.
During training, we also noticed that increasing $\tau > 0.05$ was effecting negatively to the convergence of the models. Therefore, we trained all models with $\tau = 0.05$, since it produced better results for our models.
All our results we report refer to this value of $\tau$.
We now proceed to discuss the results of our comparison between our approach and the baselines discussed in Section \ref{sec:baselines} in the two considered collective formation domains.

\subsubsection{Ridesharing}

Table \ref{rs_results} reports the results of our experiments on the ridesharing domain. Our results show that the optimality ratio obtained by our approach is comparable to the one obtained by the best performing MCTS approach (i.e., the greedy one), and clearly superior to the other MCTS approaches (including R-MCTS, the only true general approach in our comparison), who cannot compute a solution of acceptable quality. Our approach is comparable with the state-of-the-art PG\textsuperscript{2} for $n=50$ and $100$, but PG\textsuperscript{2} still outperforms our model for $n=200$. This result is not surprising, since PG\textsuperscript{2} is the state-of-the-art specifically designed for this problem domain.

\setlength\tabcolsep{5pt}
\begin{table}[]
\centering
\begin{tabular}{@{}lrrrrr@{}}
\toprule
$n$ & AM (ours) & G-MCTS & A-MCTS & R-MCTS & PG\textsuperscript{2} \\ \midrule
$50$ & $0.89$ & $0.92$ & $0.09$ & $0.08$ & $0.98$ \\
$100$ & $0.76$ & $0.88$ & $0.04$ & $0.04$ & $0.98$\\
$200$* & $0.65$ & $<0.01$ & $0.01$ & $0.02$ & $1.00$\\ \bottomrule
\end{tabular}
\caption{Optimality ratio of the attention model and baselines for a time budget of 60s for Ridesharing.
*For $n = 200$ we report the ratio with respect to the solution computed by PG\textsuperscript{2} since computing the optimal solution is not possible.}
\label{rs_results}
\end{table}

\subsubsection{Team Formation}

Table \ref{tf_results} reports the results of our experiments on the team formation domain. The optimality ratio obtained in that case is significantly better than the one obtained by other MCTS approaches. Moreover, by comparing our approach to SynTeam we can see that the gap between our approach and the domain-specific state-of-the-art approach for team formation is smaller than for ridesharing, specially for the smaller problem instances.
Notice that our approach clearly outperforms all MCTS approaches on team formation, even the one we specifically adapted for this domain (i.e., A-MCTS).

\begin{table}[]
\centering
\begin{tabular}{@{}lrrrrr@{}}
\toprule
$n$ & AM (ours) & G-MCTS & A-MCTS & R-MCTS & ST \\ \midrule
$50$ & $0.97$ & $<0.01$ & $0.84$ & $<0.01$ & $0.99$ \\
$60$ & $0.95$ & $<0.01$ & $0.78$ & $<0.01$ & $0.99$ \\
$100$* & $0.92$ & $<0.01$ & $<0.01$ & $<0.01$ & $1.00$ \\ \bottomrule
\end{tabular}
\caption{Optimality ratio of the attention model and baselines for a time budget of 60s for Team Formation. *For $n = 100$ we report the ratio with respect to the solution computed by SynTeam, since computing the optimal is not possible.}
\label{tf_results}
\end{table}

\section{Conclusions}

In this work we proposed a general approach for the formation of collectives in real-world domains based on the novel combination of an attention model and an ILP. We show that our approach is superior to previous general approaches for the formation of collectives, despite domain-specific approaches still show superior performance in some settings. 

Closing the gap with respect to domain-specific approaches is an important direction for future research. We believe that investigating alternatives to introduce diversity in the generation of collectives is a promising line of research, since we accredit entropy for a big part of the success of the attention model as an heuristic for the formation of collectives. Moreover, generalizing to larger problem instances is something our model does not excel at. Addressing these problems in future work is an important step to release the potential of our method, which is already competitive with respect to other general approaches for the generation of collectives.

Overall, we believe that this work is a first important step to foster the use of machine learning approaches for the formation of collectives. The good results achieved in the two structurally different domains that we used as benchmarks show that our attention model can indeed learn the inherent structure of the domain and exploit this to generate solutions of high quality.


\bibliographystyle{named}
\bibliography{ijcai22}

\end{document}